\documentclass[10pt,twocolumn,letterpaper]{article}

\usepackage{cvpr}
\usepackage{times}
\usepackage{epsfig}
\usepackage{graphicx}
\usepackage{amsmath}
\usepackage{amssymb}
\usepackage{mathtools}
\usepackage{float}

\usepackage[pagebackref=true,breaklinks=true,letterpaper=true,colorlinks,bookmarks=false]{hyperref}

\DeclarePairedDelimiterX{\infdivx}[2]{(}{)}{%
  #1\;\delimsize|\delimsize|\;#2%
}
\newcommand{\kld}[2]{\ensuremath{D_{KL}\infdivx{#1}{#2}}\xspace}
\newcommand\numberthis{\addtocounter{equation}{1}\tag{\theequation}}

\cvprfinalcopy %

\def\cvprPaperID{8722} %

\DeclareMathOperator*{\argmin}{arg\,min}

\graphicspath{{./figures/}}

\ifcvprfinal\fi

\begin{document}

\title{Learning to Evaluate Perception Models Using Planner-Centric Metrics}

\author{Jonah Philion\qquad Amlan Kar\qquad Sanja Fidler\\
NVIDIA\qquad University of Toronto\qquad Vector Institute\\
{\tt\small {\tt\small \{jphilion, amlank, sfidler\}@nvidia}}
}

\maketitle

\begin{abstract}
   Variants of accuracy and precision are the gold-standard by which the computer vision community measures progress of perception algorithms. One reason for the ubiquity of these metrics is that they are largely task-agnostic; we in general seek to detect zero false negatives or positives. The downside of these metrics is that, at worst, they penalize all incorrect detections equally without conditioning on the task or scene, and at best, heuristics need to be chosen to ensure that different mistakes count differently. In this paper, we propose a principled metric for 3D object detection specifically for the task of self-driving. The core idea behind our metric is to isolate the task of object detection and measure the impact the produced detections would induce on the downstream task of driving. %
   Without hand-designing it to, we find that our metric penalizes many of the mistakes that other metrics penalize by design. In addition, our metric downweighs detections based on additional factors such as distance from a detection to the ego car and the speed of the detection in intuitive ways that other detection metrics do not. For human evaluation, we generate scenes in which standard metrics and our metric disagree %
  and find that humans side with our metric 79\% of the time. 
  Our project page including an evaluation server can be found at \href{https://nv-tlabs.github.io/detection-relevance}{https://nv-tlabs.github.io/detection-relevance}.
\end{abstract}

\vspace{-4mm}
\section{Introduction}
\label{s:intro}

\begin{figure}[t]
   \begin{center}
      \includegraphics[width=1\linewidth]{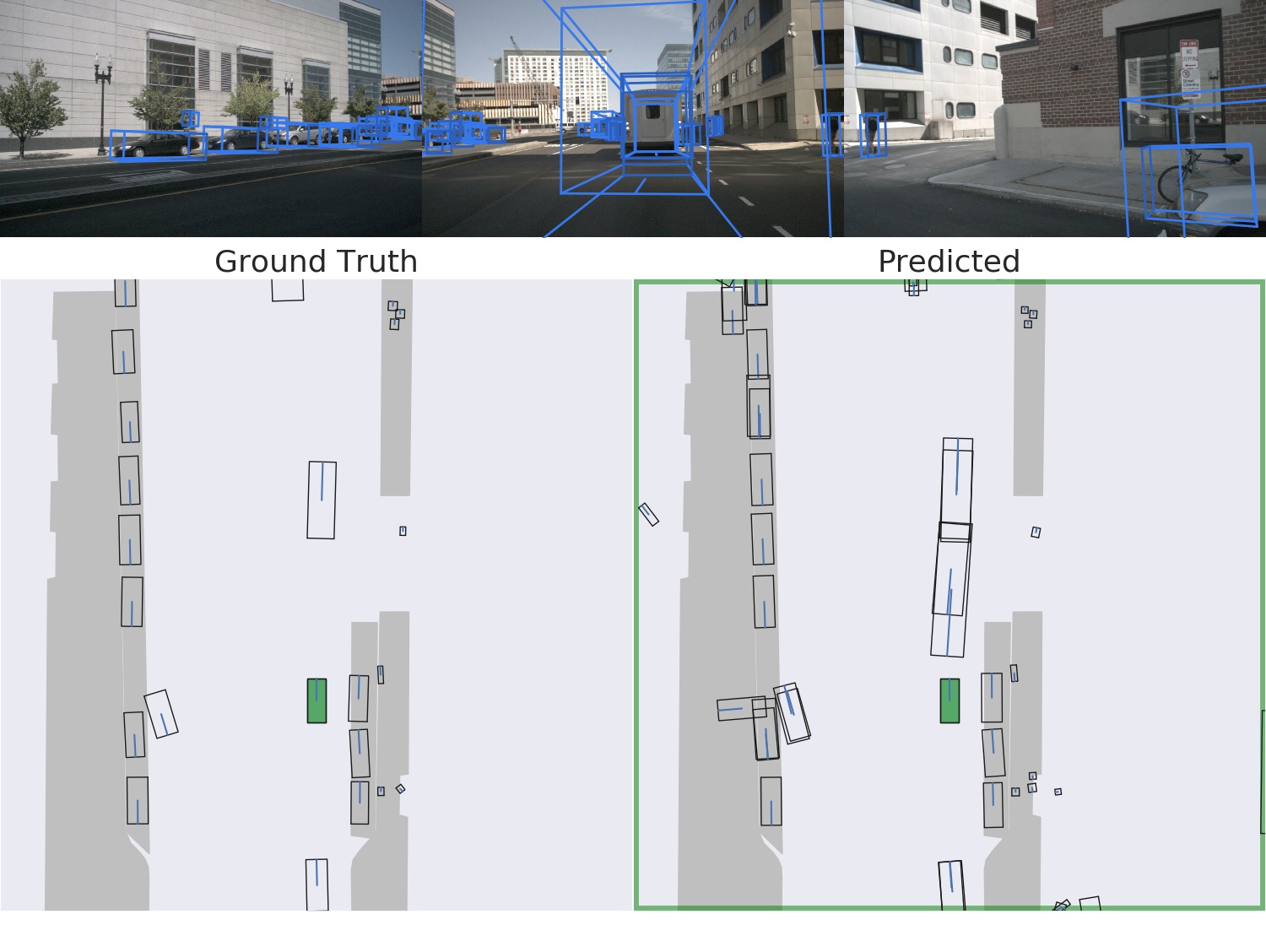}\\
   \end{center}
   \vspace{-5mm}
      \caption{\footnotesize {\bf Not all mistakes are created equal} A falsely detected parked vehicle will not lead to dangerous maneuvers by the self-driving car, while a false positive in front of it will. Metrics such as mAP penalize both cases equally. %
      Instead of hand-designing the error functions that we intuitively believe should be important for the downstream task of self-driving, we use a neural planner to rank object detectors for us. Our metric ranks the above example as the worst detection made by the state-of-the-art 3D object detector MEGVII \cite{megvii} on the validation set of nuScenes~\cite{nuscenes}.}
   \label{fig:worst}
   \vspace{-3mm}   
\end{figure}

In the past, raw accuracy and precision sufficed as canonical evaluation metrics for measuring progress in computer vision. Today, researchers should additionally try to evaluate their models along other dimensions such as robustness \cite{robustness}, speed~\cite{vidmetric}, and fairness \cite{fairness}, to name a few.
In real robotics systems such as self-driving, it is critical that perception algorithms be ranked according to their ability to  enable the downstream task of driving. An object detector that achieves higher accuracy and precision on a dataset is not guaranteed to lead to safer driving. For example, failing to detect a parked car far away in the distance, spanning perhaps only a few pixels in an image or a single LIDAR point, is considered equally bad as failing to detect a car slamming the breaks  just in front of the ego-car. 
 Ideally, our perception-evaluation metrics would more accurately translate to the real downstream driving performance. 

One way to evaluate performance is by evaluating the complete driving system either by having it drive in the real world or in simulation. Collecting real data is surely cumbersome and time consuming: since the systems are getting increasingly good, one needs to collect statistics over a very large pool of driven miles in order to get an accurate measurement. Even so, the scenarios the autonomous driving car finds itself in vary each time, and typically it is the very sparse edge cases that lead to failures.  Repeatability in the real world is thus a major issue which may lead to noisy estimates. An alternative of course is to build a perfect driving simulator in which we could sample realistic and challenging scenes and measure how different detectors affect collision rates, driving smoothness, time to destination, and other high level metrics that self-driving systems are designed to optimize for as a whole. Although progress has been made in this direction~\cite{metasim, carla, drivesim}, these simulators currently can only provide biased estimates of real-world performance.

In this paper, we propose a new metric (PKL) for 3D object detection that aligns analysis of perception performance with performance on the downstream task of driving. The key idea behind PKL is to evaluate detections through a robust planner that is trained to plan a driving trajectory based on its semantic observations, \ie, detections. By design, PKL returns the optimal score if the perception system is perfect. 
We analyze the behavior of PKL on the nuScenes dataset~\cite{nuscenes}. We show that PKL induces an intuitive ranking of the importance of detecting each vehicle in a scene. In a human study, our metric is significantly preferred over the standard metrics, even those carefully manually designed for driving~\cite{nuscenes}. To inspire the development of future perception algorithms more in line with the real-world requirements of autonomous driving, we provide a server for evaluating competing object detectors using planning-based metrics.

\section{Related Work}
\label{s:related}
\textbf{Evaluation Metrics:} Evaluation of trained neural networks is an active area of research. Most recently, ``average delay'' \cite{vidmetric} has been proposed as an alternative to average precision for object detectors that operate on videos. In the field of autonomous vehicles, metrics such as nuScenes Detection Score \cite{nuscenes} and ``Mean average precision weighted by heading'' \cite{waymo_open_dataset} have been proposed as metrics that rank detectors with hand-crafted penalties that align with human notions of safe driving. Our goal in this paper is to train a planning network that can learn what aspects of detection are important for the driving task, then use this network to measure performance of upstream detectors.

\textbf{3D Object Detection:} The task of 3D object detection is to identify all objects in a scene as well their 6 degree-of-freedom pose. Unlike lane detection or SLAM which can be bootstrapped by high-definition maps and GPS, 3D object detection relies heavily on realtime computer vision. As a result, recent industrial-grade datasets largely focus on solving the 3D object detection problem \cite{nuscenes, argoverse, lyft2019, waymo_open_dataset}.

Contemporary object detectors are largely characterized by the kind of data that they take as input. Among detectors that only take LiDAR as input, PointPillars \cite{pointpillars, second}, and PIXOR \cite{pixor} represent two variants of architectures; models based on PointPillars apply a shallow PointNet \cite{pointnet} in their first layer while models based on PIXOR discretize the height dimension \cite{megvii, sparse_to_dense, segheight}. Camera-only 3D object detectors either use 3D anchors that are projected into the camera plane \cite{OFT, Chen_2016_CVPR} or use separate depth prediction networks to lift 2d object detections in the image plane to 3D \cite{mair}. Approaches that attempt to use both LiDAR and camera modalities \cite{Liang_2019_CVPR} have lacked in performance what they possess in complexity. Across all data modalities, these approaches are ranked according to mean average precision over a set of hand-picked distance thresholds and measures of object visibility \cite{kitti, nuscenes}.

\textbf{End-to-end Planning:} End-to-end driving is a tantalizingly scalable solution to the self-driving problem. Recent work in self-driving has focused on modeling the driving problem so that the entire system can be optimized through gradient descent \cite{nvendtoend, driveinaday}. ChauffeurNet \cite{chauffeurnet} trains agents on large amounts of data to autoregressively generate future trajectories given perception output. PRECOG \cite{precog} conditions on LiDAR point clouds to generate a joint distribution over future trajectories for all agents in the scene. Neural Motion Planner \cite{neuralmotionplanner} also uses teacher trajectories to learn a distribution over trajectories but uses a hard-margin loss that includes other priors on behavior such as traffic rules. While end-to-end approaches that operate directly on raw sensor inputs are highly scalable, Zhou et al. \cite{koltun_action} suggests that explicit perception bottlenecks result in better performance on the downstream tasks.

\section{Methodology}
\label{s:method}
In this section, we motivate the definition of our PKL metric. While the vast majority of evaluation metrics are analytic, our metric requires a preliminary optimization. We explain how we parameterize the metric and how we learn the parameters from data.

\begin{figure}[t] 
\begin{center}
   \includegraphics[width=0.95\linewidth]{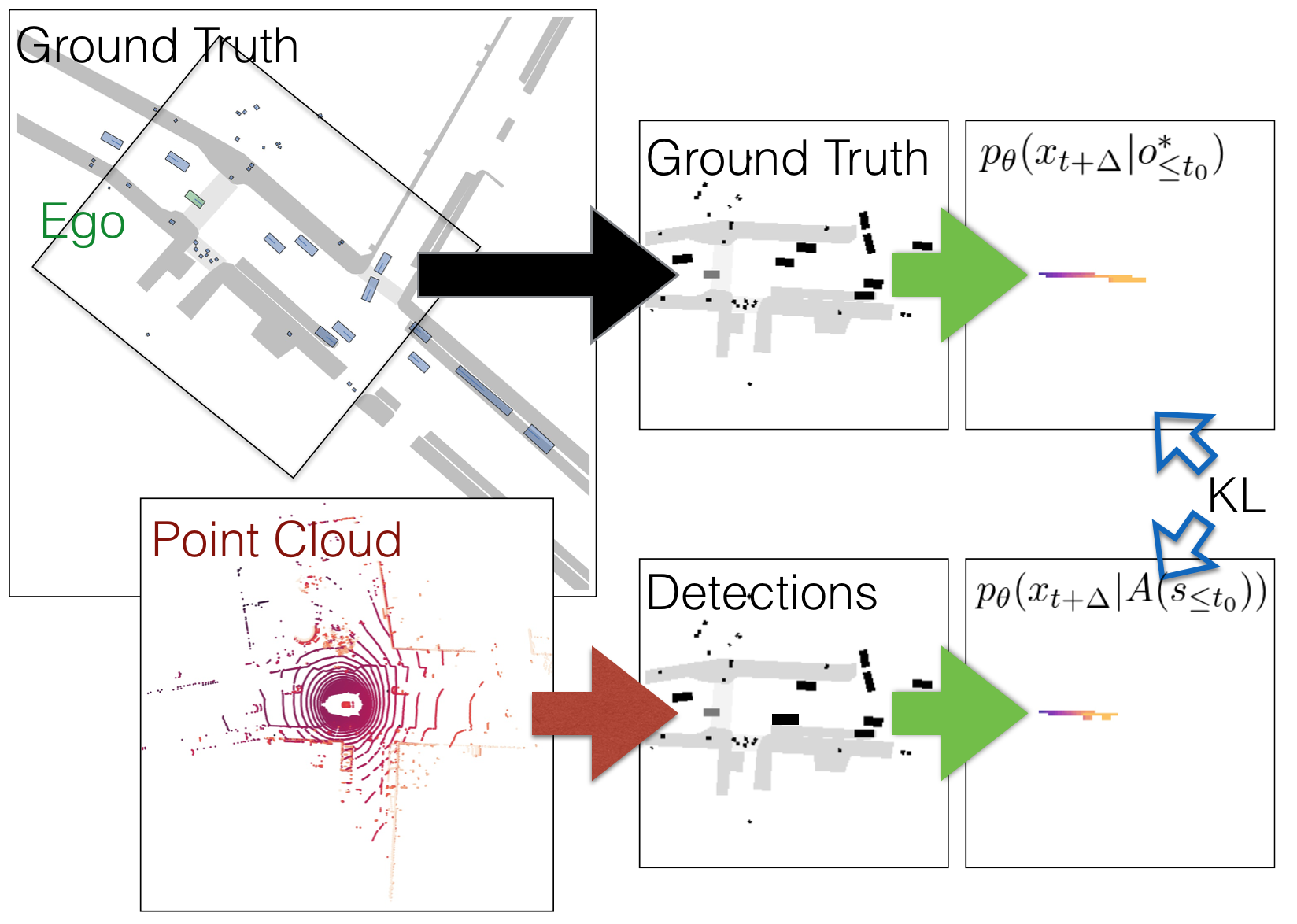}
\end{center}
\vspace{-3mm}
   \caption{{\bf PKL} We model $p_\theta(x_t | o_{\leq t})$ in the local frame of each vehicle with a CNN (green). $o_{\leq t}$ includes all map data and detected objects from the previous 2 seconds. For a detector $A$ (red), our metric is defined by PKL($A$)=$\kld{p_\theta(x_t | o^*_{\leq t})}{p_\theta(x_t | A(s_{\leq t}))}$ where $s_t$ includes sensor modalities that the object detector $A$ requires and $o^*$ includes ground truth detections. If the detector is perfect, the PKL is 0. See Section~\ref{ss:pkl} for details.}
\label{fig:intro}
\vspace{-2mm}
\end{figure}

\subsection{Background}
\label{ss:pkl}
We wish to measure how the future state of a multi-agent system operating under some dynamics 
changes due to a noisy agent, which is our self-driving car. For the purpose of 
measuring perception performance, we consider that the noise in our agent comes only from
noisy perception. Let $x_t^i$ denote the position of agent $i \in \{1 \ldots N\}$ at time $t$ and $o_t^{i}$ denote the observation (coming from perception) for agent $i$ at time $t$. We will denote the perfect
perceptual observation as $o_t^{i*}$. The joint probability of the ``ideal'' system state over a time horizon of $T$ time steps starting from $t=1$ is,
\begin{align*}
   P = p(x^1_1 \ldots x^N_T | o^{*1}_1 \ldots o^{*N}_T) \numberthis
\end{align*}

Without loss of generality, we will consider the first agent to be our noisy agent. The metric we want to compute is the change in this distribution given noisy observations from our agent, which can be measured using the KL Divergence \cite{kldiv} as follows,
\begin{align*}
   &\kld{P}{Q} \numberthis \\
   &P = p(x^1_1 \ldots x^N_T | o^{*1}_1 \ldots o^{*N}_T) \\
   &Q = p(x^1_1 \ldots x^N_T | o^1_1 \ldots o^1_T, o^{*2}_1 \ldots o^{*N}_T)
\end{align*}

To measure the perception performance at $t=1$, we first assume that all agents make future predictions given observations only at $t=1$. We discuss this assumption in detail at the end of this section. The joint probability can then be written as,
\begin{align*}
   P = p(x^1_1 \ldots x^N_T | o^{*1}_1 \ldots o^{*N}_1) \numberthis
\end{align*}

Since the agents do not get any new observations in this time horizon of $T$ steps, they can only act independently of each other (since their future states are not observable to each other). The joint probability then becomes a product of the marginal distributions over the future of every agent,
\begin{align*}
   P = \prod_{i=1}^N p(x^i_1 \ldots x^i_T | o^{*i}_1) \numberthis
\end{align*}

Finally, we assume that the system moves independently at each time step, given its observations. This amounts to factorizing the joint probability as,
\begin{align*}
   P = \prod_{t=1}^T \prod_{i=1}^N p(x^i_t | o^{*i}_1) \numberthis
\end{align*}

Under these assumptions, the joint distribution Q under noisy observations from our agent factorizes as,
\begin{align*}
   Q = \prod_{t=1}^T p(x^1_t | o^{1}_1) \prod_{i=2}^N p(x^i_t | o^{*i}_1) \numberthis
\end{align*}

Substituting these in the KL divergence, we get,
\begin{align*}
   &\kld{P}{Q} \\
   &= \mathbb{E}_P \bigg[ \log \frac{\prod_{t=1}^T \prod_{i=1}^N p(x^i_t | o^{*i}_1)}{\prod_{t=1}^T p(x^1_t | o^{1}_1) \prod_{i=2}^N p(x^i_t | o^{*i}_1)} \bigg] \numberthis\\
   &= \mathbb{E}_P \bigg[ \log \frac{\prod_{t=1}^T p(x^1_t | o^{*1}_1)}{\prod_{t=1}^T p(x^1_t | o^{1}_1)} \bigg] \numberthis\\
   &=\kld{P^1}{Q^1} \numberthis
\end{align*}
where, $P^1$, $Q^1$ represent the marginal distribution over the future states of our agent, given perfect and noisy perception, respectively. In practice, these assumptions make computing the metric tractable, since we can train a parametric model of possible future states of an agent $p_\theta(x_t | \mathbf{o})$. The specific instantiation of state $x_t$, observations $\mathbf{o}$, model $p_\theta$ and its training is presented in Section~\ref{ss:planner}. 

\textbf{Discussion on assumptions:} To obtain a tractable estimate of the metric, and to measure the performance of perception at a particular time $t$, we assumed that predictions over a time horizon $T$ from $t$ are made given only the initial observation at $t$, and that every agent acts independently of each other and at every time step in this time. The first assumption is the most important and
entails that all agents in the scene are not ``reactive'' in the time horizon specified by $T$. This enables us to measure how well perception till (or at) a current time step can help in driving with ``anticipation''. Within a short time horizon $T$, this is indeed intuitive, since perfect perception should result in a best-case scenario for anticipatory driving. This is reflected in the PKL metric, which is zero when $o_1^{*1} = o_1^{1}$. Moreover, imperfect perception in irrelevant parts of a scene, such as in a nearby parking lot will not affect the metric since it does not affect how the whole system would have progressed in time. The second assumption follows from the first, since given no new sensory information, the agents can only act independently of each other. The last assumption is not necessary to our derivation, but is used in our particular implementation -- where we model the marginal likelihood of an agent's location at every time step within the time horizon $T$ independently, as explained in Section~\ref{ss:planner}.

\subsection{``Planning KL-Divergence (PKL)''}
\label{ss:planner}
Let $s_1,...,s_t \in S$ be a sequence of raw sensor observations, $o^*_1,...,o_t^* \in O$ be the corresponding sequence of ground truth object detections, and $x_1,...,x_t$ be the corresponding sequence of poses of the ego vehicle. Let $A: S \rightarrow O$ be an object detector that predicts $o_t$ conditioned on $s_t$. We define the PKL at time $t$ as
\begin{align*}
   \mathrm{PKL}&(A) \numberthis\\
   &= \sum_{0<\Delta\leq T} \kld{p_\theta(x_{t+\Delta} | o^*_{\leq t_0})}{p_\theta(x_{t+\Delta} | A(s_{\leq t_0}))}
\end{align*}
where $p_\theta(x_t | o_{\leq t})$ models the distribution of ground truth trajectories in the dataset $D$,
\begin{align*}
   \theta = \argmin_{\theta'} \sum_{x_t \in D} -\log p_{\theta'} (x_t | o^*_{\leq t}). \numberthis
\end{align*}

Intuitively, the PKL is a way to measure how similar a set of detections in a scene are from the ground truth detections. It does so by measuring how differently the ego car would plan if it only saw the predicted objects versus seeing the actual objects in the scene.

We model the marginal likelihoods of future positions with a similar approach to other end-to-end planning architectures \cite{neuralmotionplanner,chauffeurnet}. We discretize the grid -17.0 meters behind the ego to 60.0 meters in front of the ego and $\pm 38.5$ on either side into voxels of size 0.3 meters by 0.3 meters. We form the input $x \in \mathbb{R}^{8 \times X \times Y}$ by binarizing the 3 map layers ``ped\_crossing'', ``walkway'', and ``carpark\_area'' and concatenating with binarized birds-eye-view projections of the detections for $t \in \{ t_0 - 2.0, t_0 - 1.5, t_0 - 1.0, t_0 - 0.5, t_0 \}$ where all coordinates are transformed to the frame of the ego car from time $t_0$. To form the target, we discretize the ground truth trajectory of the ego for timesteps $\{ t_0 + 0.25 i \mid 0<i<16, i \in \mathbb{N}\}$ and train with cross entropy loss as is standard for segmentation. We train using all non-zero trajectories of all annotated cars in nuScenes training set (1,216,412 trajectories) with batch size 16 for 100k steps using Adam \cite{adam,optimsearch} with learning rate 2e-3 and weight decay 1e-5. We validate only on ego trajectories from the validation set (4,135 trajectories). To find the PKL over the full dataset, we average over all 2 second chunks. Note this is one possible  instantiation of a neural planner, and other parametrizations and designs are possible. Our key contribution is in exploiting  (neural) planner in evaluating perception models.

\section{Experiments}
\label{s:experiments}

While we make no claim that the conditional generative model of trajectories trained using the protocol described above is perfect, we seek to demonstrate empirically that the model is ``good enough'' in the sense that aspects of detection that are intuitively salient for the self-driving task are reflected in the distributions output by the planning model and humans generally side with detection rankings induced by PKL over other metrics.

We validate our proposed evaluation metric on the nuScenes dataset~\cite{nuscenes}. nuScenes consists of 1000 annotated driving scenes each of length 20 seconds, that are taken from busy local roads in Boston and Singapore. Ground truth 3D object labels are provided at 2 hz for objects that fall into 10 object classes including cars, trucks, pedestrians, and road barriers. The dataset contains 1.4M camera images, 390k LIDAR sweeps, 1.4M RADAR sweeps, and 7x more object labels than KITTI~\cite{kitti}.

\begin{figure}[t!]
\vspace{-2mm}
\begin{center}
   \includegraphics[width=0.95\linewidth]{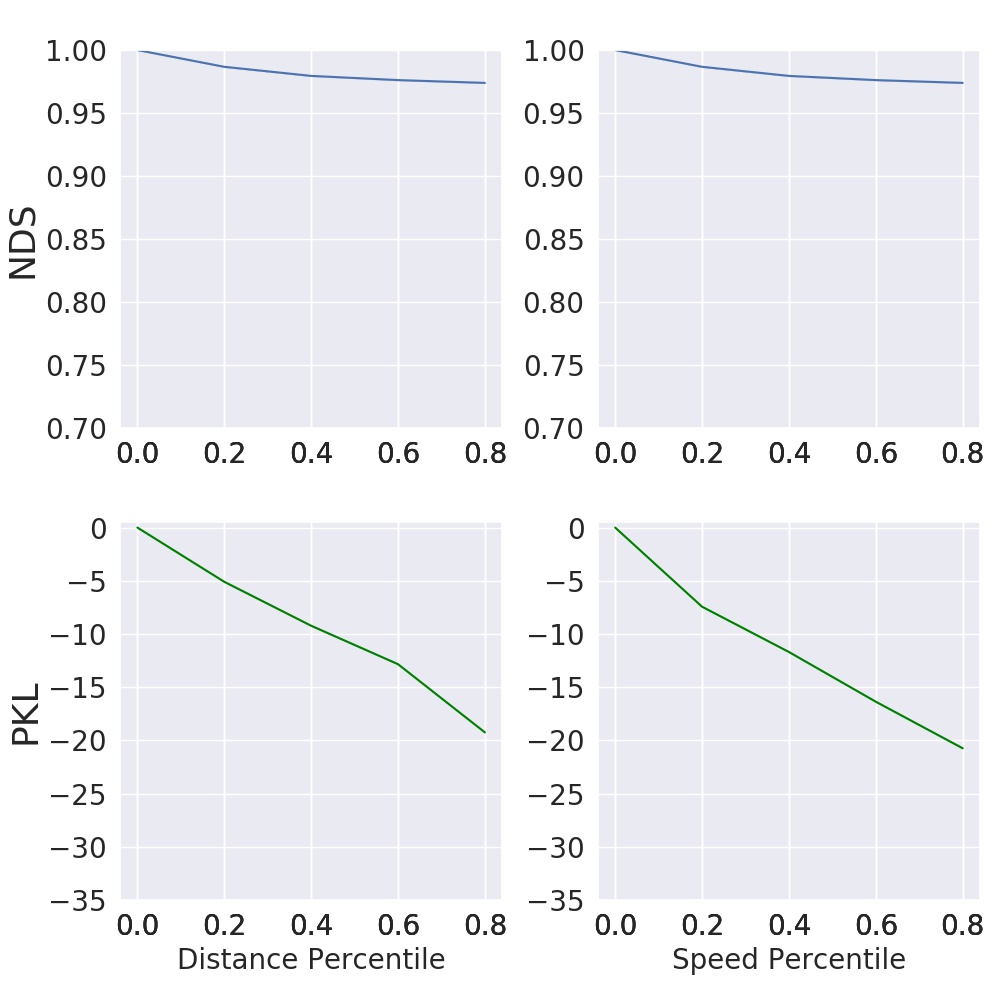}
\end{center}
   \vspace{-4mm}
   \caption{{\bf PKL takes into account context, unlike NDS} The carefully manually designed NDS metric~\cite{nuscenes} (left) is largely invariant to the location and speed of the objects that the object detector misses. PKL on the other hand penalizes missed detections of faster moving vehicles that are closer to the ego car. PKL is consistent with human intuition on which objects are most critical for safe driving as supported by Table~\ref{tab:human}.}
\label{fig:antialign}
\vspace{-4mm}
\end{figure}

\begin{figure*}[t!]
\vspace{-2mm}
   \begin{center}
      \includegraphics[width=0.95\linewidth]{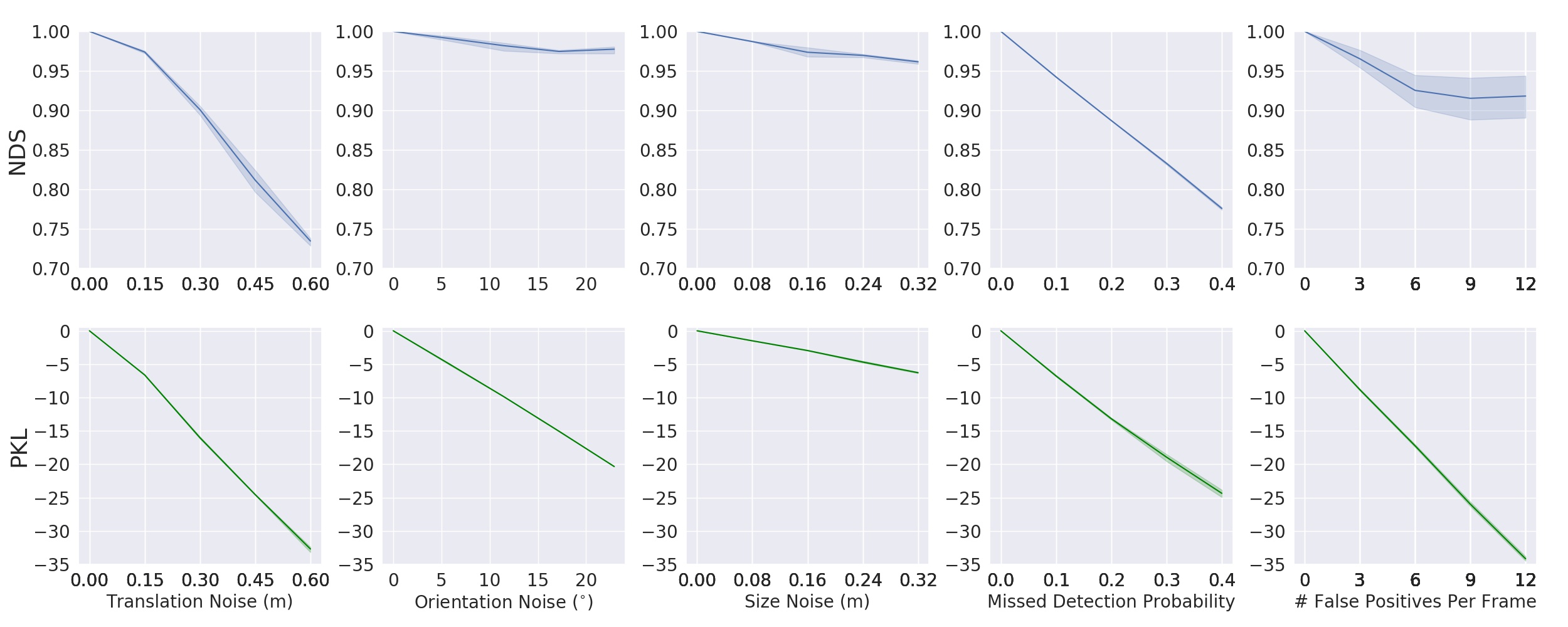}
   \end{center}
   \vspace{-3mm}
      \caption{{\bf PKL and NDS are correlated under certain noise models} We add synthetic noise to the ground truth detections in the dataset and observe how the noise affects the nuScenes Detection Score (NDS) \cite{nuscenes} and PKL. We find that NDS and PKL are tightly correlated across noise models. ``Translation noise'', ``Orientation noise'', and ``Size noise'' refer to adding gaussians with increasing variance to the ground truth labels. For ``Missed Detection Probability'', we drop detections with probability $p$. ``False positives'' are generated by placing cars uniformly randomly within a bounding box of the ego car (Sec.~\ref{ss:metric_align}). While NDS is engineered to be negatively correlated with these quantities, these correlations arise from PKL because of the affect they have on the downstream planning task.}
   \label{fig:align}
   \end{figure*}

\begin{table}[h!]
   \begin{center}
   \begin{tabular}{|l|c|c|c|c|}
   \hline
   Method & Top 5 & Top 1 & $||x_{gt} - x_{pred}||$ \\
   \hline\hline
   Ours & \textbf{37.41}\% & \textbf{19.39}\% & \textbf{1.27} m\\
   -ego only & 35.28 & 17.18 & 1.47\\
   -loss clip \cite{detectionfreebies} & 35.72 & 18.33 & 1.45\\
   -pos weight & 35.47 & 18.74 & 1.42\\
   -dropout \cite{dropout} & 34.25 & 18.59 & 1.49\\
   \hline
   \end{tabular}
   \end{center}
      \vspace{-2mm}
   \caption{{\bf Planner performance} Dropout, loss clipping, and loss function weighting are techniques for fighting class imbalance and overfitting. We show that on nuScenes val, the combination of these techniques along with treating labeled objects as ego vehicles results in the best Top 5 accuracy, top 1 accuracy, and L2 distance between the mode of the predicted distribution and the ground truth future position. Importantly, we only measure these quantities for the ego car trajectories during evaluation independent of training hyperparameters.}
   \vspace{-2mm}
   \label{tab:planablation}
   \end{table}

\subsection{Planner Ablation} 
\label{ss:ablate}

We present a short analysis on the planner's performance w.r.t. different training hyperparameters in Tab.~\ref{tab:planablation}. Due to class imbalance, we find that weighting positive examples and clipping the loss function \cite{detectionfreebies} provides accuracy boosts. Although we report accuracies exclusively on ego vehicle drives from the validation set, we find that training on the trajectories of all annotated vehicles in the dataset results in the largest boost to performance. We measure Top $k$ accuracy by calculating the Top $k$ locations in each heat map $p(x_{t+\delta} | o)$ and averaging over $\delta$ and $t$.
The more accurately the planner is able to approximate the distribution of feasible future trajectories, the better the ranking induced by the planner will be.

\begin{figure}[t!]
   \vspace{-2mm}
   \begin{center}
      \includegraphics[width=1\linewidth]{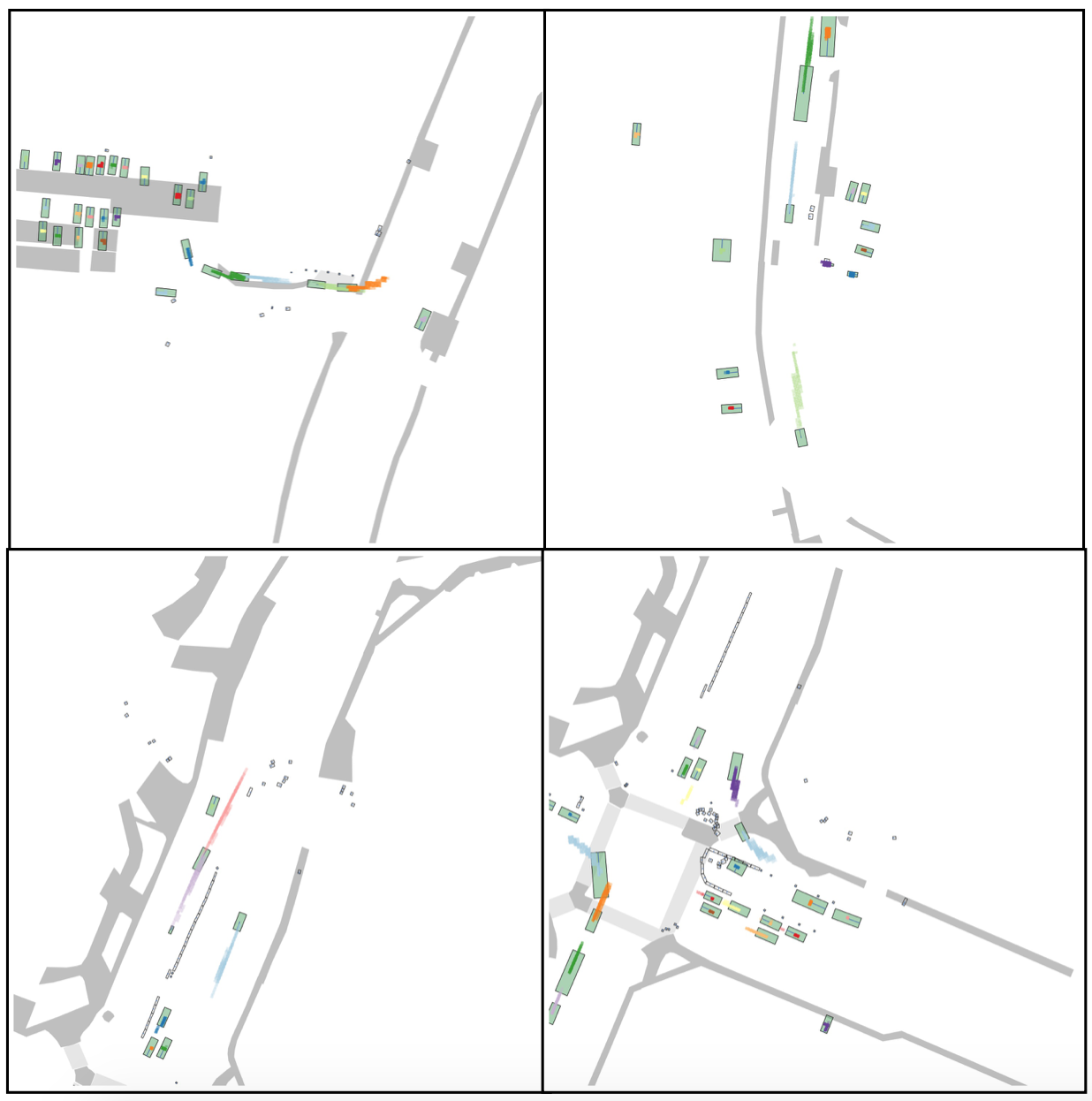}
   \end{center}
   \vspace{-3mm}
      \caption{{\bf Trajectory heatmap visualization} Because we train on all labeled vehicles in the training set, our planner is in theory capable of forecasting in the frame of any detected vehicle in the validation set. For simplicity, we visualize the heatmaps for all future timesteps as a single color with varying transparency. Different objects are given one of ten different colors to facilitate matching between cars and heatmaps.}
   \label{fig:forecast}
   \end{figure}

We qualitatively demonstrate our planner in Figure~\ref{fig:forecast}. We sample frames from the validation set and visualize the planner's predictions for all vehicles that have existed for longer than 2 seconds in the current frame. More examples can be found on the \href{https://nv-tlabs.github.io/detection-relevance/}{project page}.

\subsection{Aligning with Existing Metrics}
\label{ss:metric_align}
The nuScenes object detection benchmark uses a heavily engineered evaluation metric, called the nuScenes Dataset Score (NDS) to rank object detectors~\cite{nuscenes}. NDS is defined as:
\begin{align*}
   &\text{NDS} \numberthis\\
   &= \frac{1}{2}\left[\text{mAP} + \frac{1}{|TP|}\sum_{mTP \in TP}(1 - \min(1, mTP))\right]
\end{align*}
where $TP$ is a collection of ``true positive'' error functions that are only measured on detections that are matched with a ground truth detection. NDS is designed to penalize false positives, false negatives, orientation error and translation error for all ground truth boxes within a distance $d_k$ to the ego car for each class of object $k$. This behavior is chosen because it aligns well with human intuition on what is important to perceive in order to drive safely.
We show that our metric is also sensitive to these errors. More importantly, in our metric, these properties \emph{emerge} because the planner implicitly learns that these variables are strong signals for predicting the distribution of future trajectories.

Results are shown in Fig.~\ref{fig:align}. We show that our metric possesses these properties by evaluating NDS and PKL on detectors with synthetic noise. To test translation error, we add gaussian noise to the center coordinate of every ground truth box. To test orientation error, we add gaussian noise to the 2D heading of every ground truth box. To test size noise, we add gaussian noise to the width, length, and height of every box. To test response to false negatives, we drop every detection with some fixed probability $p$. To test response to false positives, we add $N$ boxes of random size, orientation, and location into the scene at each timestep.

We see that for all noise models, NDS and PKL decrease with more error. Interestingly, PKL penalizes orientation more strongly than NDS. In the recently released Waymo Open Dataset~\cite{waymo_open_dataset}, a new metric named ``Mean average precision weighted by heading'' or mAPH was proposed. mAPH is designed to weigh heading more heavily than the size, center of the bounding box because future prediction is generally more sensitive to the heading. We find it compelling that our metric implictly learns this weighting.

\subsection{Conditioning on Context}
\label{ss:metric_misalign}
While NDS and mAP are guaranteed to agree with intuition about the importance of detecting objects accurately, they do not condition on a specific scene to determine how important each detection is \emph{in context}. For instance, an object detector that always predicts a false positive directly in front of the ego vehicle receives roughly the same score under mAP as a detector that predicts a false positive 30 meters behind it. If the downstream task for the detector is unknown, it is difficult to justify weighing certain detections more than others. 

In Fig~\ref{fig:antialign}, we show that our metric learns to take these factors into account. In the first row of Fig.~\ref{fig:antialign}, for each scene, we remove 5 vehicles with distance in the $p$ percentile of distances among all objects in the scene. As a result, in each trial, we get roughly the same number of false negatives, but the distribution of distances of removed cars to the ego car decreases with increasing $p$. For the second row, we rank the cars by speed in the global frame instead. In this case, the distribution of speeds of removed vehicles increases with increasing $p$. Unlike the noise models visualized in Fig.~\ref{fig:align}, these noise models are deterministic so we do not display error bars. Our metric penalizes missed detections closer to the vehicle as well as missed detections that are moving at high speed. However, the NDS score stays roughly the same in this experiment. The behaviour in PKL strongly correlates with intuition, where these detections would be considered critical to safe driving.

   \subsection{MEGVII Best and Worst}
   \label{ss:bestworst}
\begin{figure}[t!]
   \vspace{-5mm}
      \begin{center}
         \includegraphics[width=0.8\linewidth]{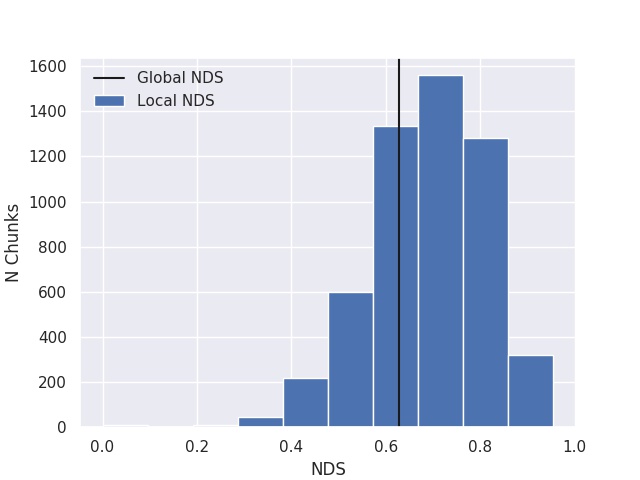}
      \end{center}
      \vspace{-3mm}
         \caption{{\bf ``Local" NDS} NDS is a global metric similar to BLEU \cite{bleu}. We show that over all of the MEGVII detections on the nuScenes validation set, our local approximation of NDS is a decent monte carlo estimate of the global NDS.}
      \label{fig:localnds}
      \end{figure}

   To gain insight into what the different metrics penalize, we rank the scenes in the dataset according to the performance of the state-of-the-art 3D object detector, MEGVII~\cite{megvii}. While ranking under PKL comes naturally given that the PKL is the expectation of KL over all scenes, NDS is not written as an expectation and therefore needs to be adapted. We adapt the NDS by calculating average precision (AP) only for classes that have ground truth boxes in each local chunk of a scene. Fig.~\ref{fig:localnds} shows that this temporally local version of NDS is a well-behaved approximation of the global NDS.
   
   Fig.~\ref{fig:worst} shows the time chunk on which the published MEGVII detections perform worst under the PKL metric. In the scene, a false positive appears right in front of the ego vehicle, giving the appearance that the truck in front of the ego is moving backwards. As a result, the planner expects the ego vehicle to stop instead of continuing forward, resulting in a huge penalty under the PKL metric.

   \begin{figure*}[t!]
      \begin{center}
      \includegraphics[width=0.95\linewidth]{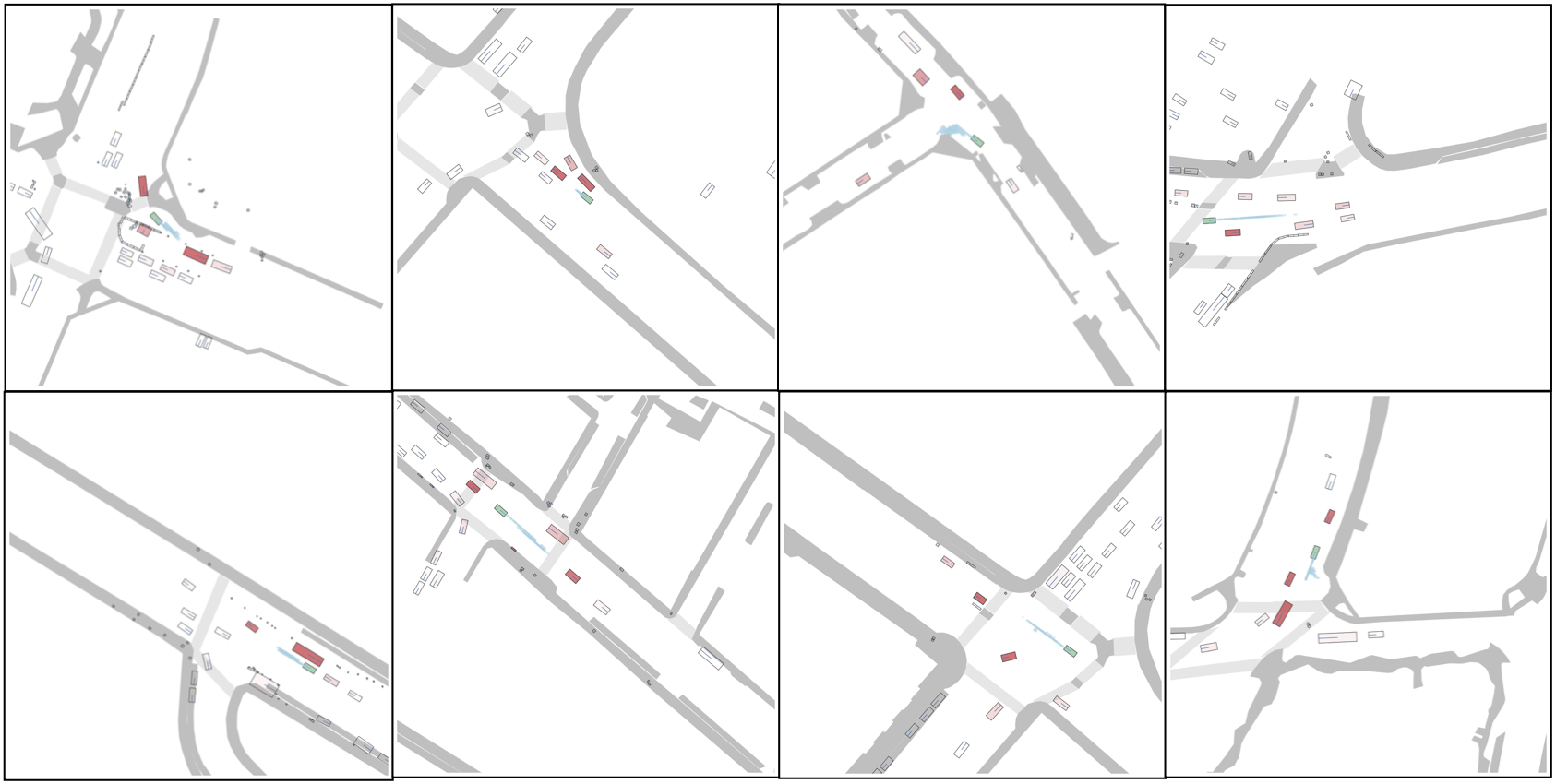}
      \end{center}
      \vspace{-3.5mm}
         \caption{{\bf False negative sensitivity} We remove each ground truth detection from a scene and evaluate the PKL. Ego car is shown in green. Detections that resulted in a larger PKL when they were removed are visualized in red. The objects found to be important are intuitive, but not necessarily the closest object to the ego-car.}
      \label{fig:importance} 
      \end{figure*}
   
   \begin{figure*}[t!]
     \vspace{-3mm}
     \begin{minipage}{0.49\linewidth}
      \begin{center}
         \includegraphics[width=0.94\linewidth]{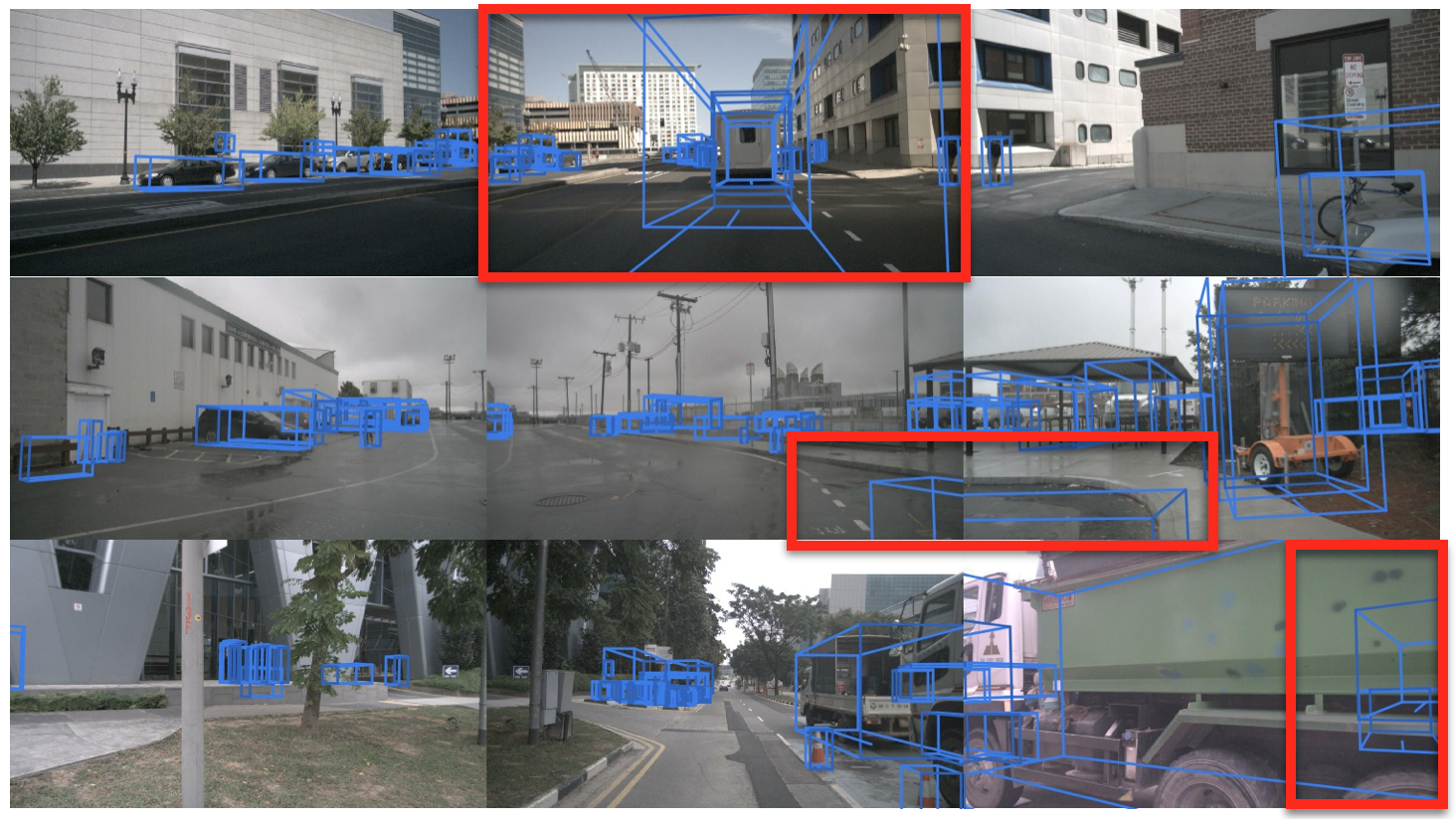}
      \end{center}
        \vspace{-3.5mm}
      \caption{\footnotesize {\bf High PKL MEGVII mistakes} MEGVII detections ranked \textit{most dangerous} under the PKL metric. Most of the bottom ranked instances include false positives that are close to the ego vehicle.}
      \label{fig:best}
  \end{minipage}
  \hspace{2mm}
   \begin{minipage}{0.49\linewidth}
   \vspace{-2mm}
      \begin{center}
         \includegraphics[width=0.94\linewidth]{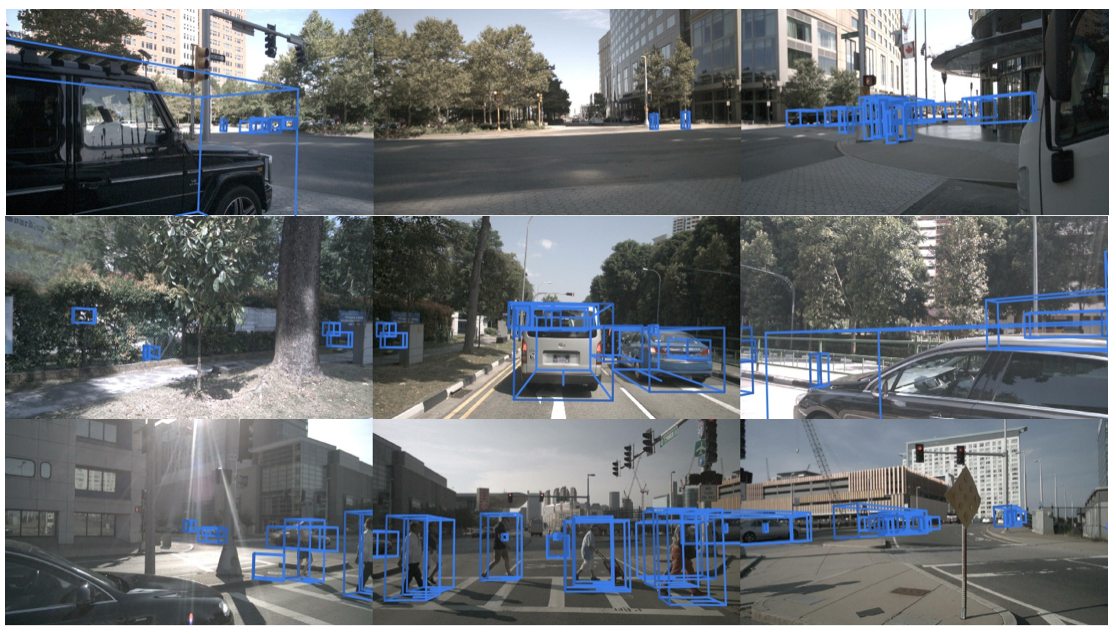}
      \end{center}
        \vspace{-3.5mm}
      \caption{{\bf Low PKL MEGVII mistakes} MEGVII detections on the nuScenes validation set ranked \textit{least dangerous} under the PKL metric.}
      \label{fig:best}
        \end{minipage}
           \vspace{-2.5mm}
   \end{figure*}
   
   \begin{table}[t!]
   \begin{center}
   \begin{tabular}{|l|c||c|c|}
   \hline
   Scenes & Responses & NDS & PKL\\
   \hline\hline
   75 & 730 & 21\% & \textbf{79\%}\\
   \hline
   \end{tabular}
   \end{center}
   \caption{{\bf Human evaluation} Humans side with PKL over NDS 79\% of the time on what kinds of detection errors are more dangerous.}
   \label{tab:human}
\end{table}

\begin{figure}[t!]
   \begin{center}
      \includegraphics[width=0.95\linewidth]{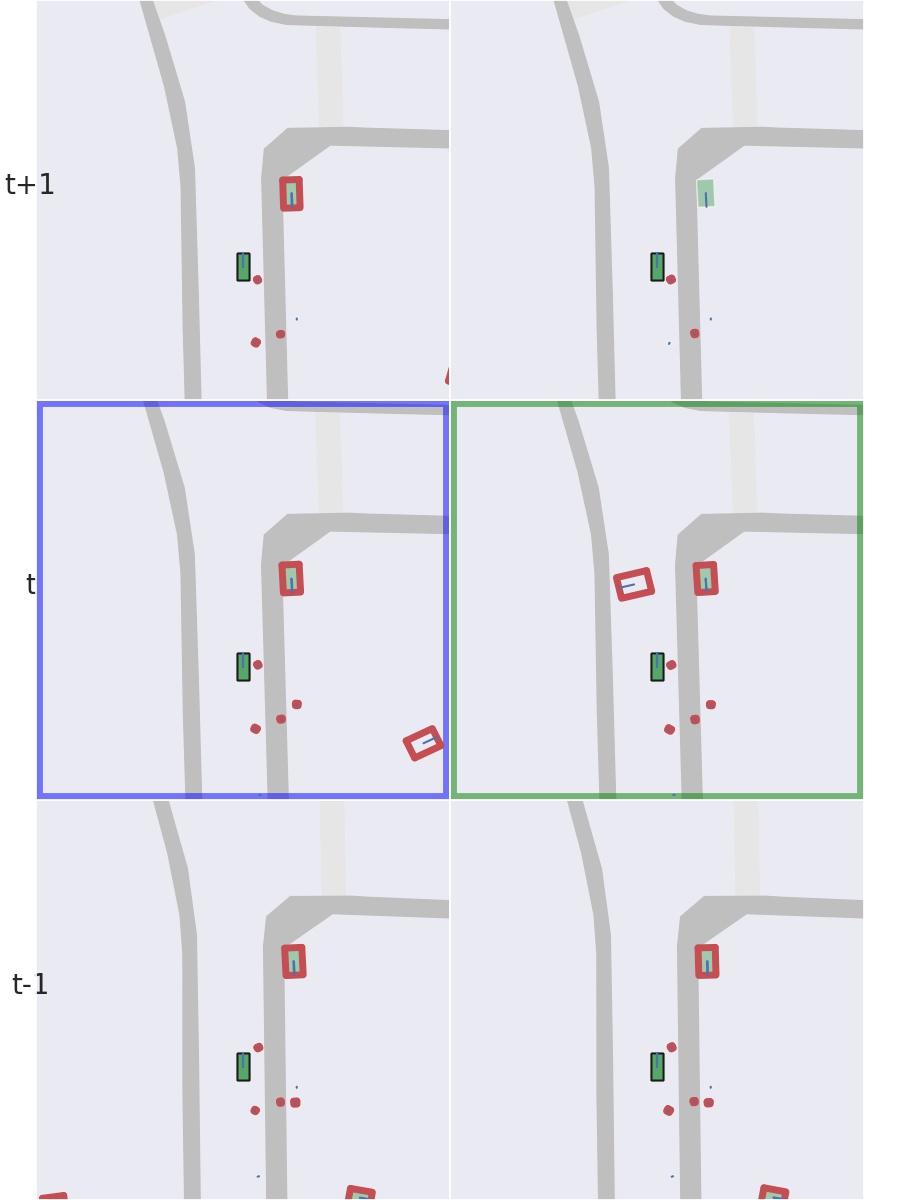}
   \end{center}
   \vspace{-3mm}
      \caption{\footnotesize {\bf AMT example} We use Amazon Mechanical Turk to test the extent to which PKL aligns with human notions of safety. We show GIFs of length 2 seconds of the same scene but with different noise models applied to the ground truth annotations. In the example above, NDS penalizes the left column more strongly than it penalizes the right, but PKL recognizes the false positive as dangerous, which also aligns with the human opinions. More examples shown to the turkers can be found on the \href{https://nv-tlabs.github.io/detection-relevance/}{project page}.}
   \label{fig:turkex}
      \vspace{-2mm}
\end{figure}

   Figure~\ref{fig:best} shows the time chunk on which MEGVII performs best under PKL. In the time series, the detection of the car to the left of the ego is stable. There are several false positive humans detected in the scene, but these detections are irrelevant to the task of waiting at the light, which is why the scene still performs well. We recognize that for some downstream tasks, such as autonomous taxis, accurately detecting the humans on the sidewalk is a crucial subtask. Our goal is not to advocate for the sole adoption of PKL to evaluate object detectors but to propose PKL as an alternative to task-agnostic metrics that do not account for the context in which perceptual mistakes are made.

\subsection{Human Evaluation}
\label{ss:human}

\begin{figure}[t!]
   \begin{center}
      \includegraphics[width=1\linewidth,trim=110 0 110 0,clip]{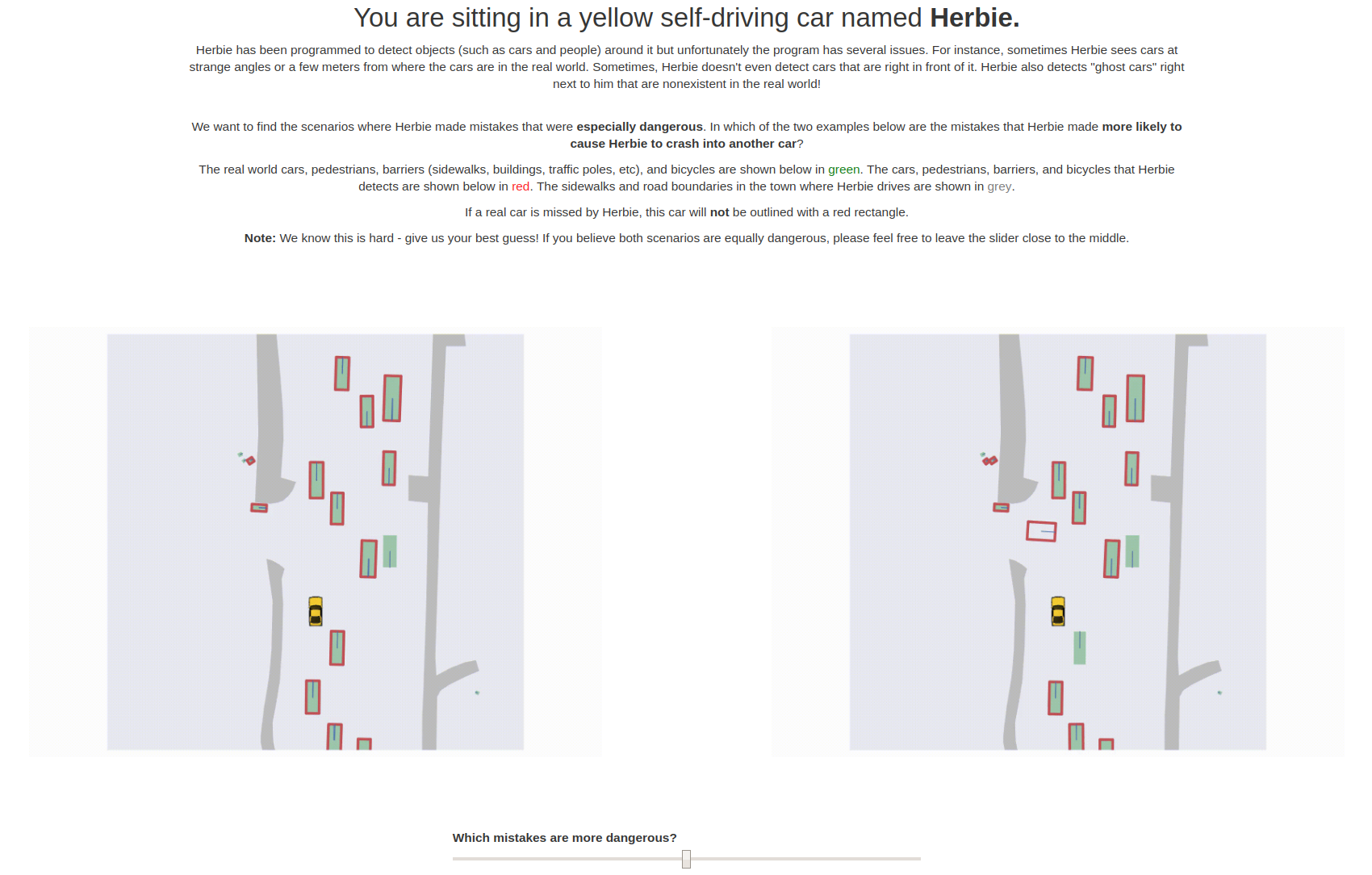}
   \end{center}
    \vspace{-2mm}
      \caption{\footnotesize {\bf AMT instructions} A screenshot from the survey that we use. Note that the driving examples are gifs in the real survey. In the above example, NDS ranks the left sequence as more dangerous but most people would agree that the false positive and negative on the right are potentially more dangerous. Instructions can also be found on the \href{https://nv-tlabs.github.io/detection-relevance/}{project page}.}
   \label{fig:mturk}
    \vspace{-2mm}
\end{figure}

We submit a survey to the Amazon Mechanical Turk service asking humans to decide if one set of noisy detections is more dangerous than another set of noisy detections in a certain scene. Instructions provided to the workers are shown in Fig~\ref{fig:mturk}. We name the car ``Herbie'' to encourage workers to empathize with the car. We choose the scenes and noise such that NDS and PKL disagree on which scene has noise that is more dangerous. Maintaining the same scene for a given pair forces workers to differentiate between the two options based purely on the behavior of the detections as opposed to differences in the complexity of the scenes. Noise is added to the system to differentiate between metrics based on how they couple across error functions; we generate noisy detections by sampling translation noise with $\sigma=0.1$m, orientation noise with $\sigma=4^\circ$, size noise with $\sigma=0.1$m, missed detection with probability $p=0.05$, and exactly 1 false positive per frame.

While PKL is defined as the expectation of PKL for a single frame, there is no obvious way to obtain monte carlo estimates of mAP for single samples. In NDS, this problem is exacerbated by the fact that the mAP is normalized over classes which would mean that scenes with very few instances of a class would be unfairly penalized. We approximate mAP by evaluating mAP over a segment in time only for classes that have at least one ground truth box within that time segment. We visualize the histogram of these local NDS measurements in Figure~\ref{fig:localnds} to verify that we can provide a competitive ranking under the local NDS metric.

We leave an optional comment box on the survey. Workers largely appear to pay attention to the correct mistakes made by the detectors. For instance, common comments include ``failure to detect vehicle behind'', ``The car to the left wasn't detected but it's off to the side'', and ``something isn't in Herbie's path but it thinks something is''. However, it is not clear that all workers fully understand the task that they are being asked to enact. Other comments include ``Herbie runs into an object'', ``looks like it thought it had a collision'', ``there looks to be a possible head on collision here'', suggesting that the concepts of false positive and false negative are not easily communicated through the survey to a naive crowd without technical expertise.

\section{Discussion}
\label{s:discussion}

Conditioned on any arrangement of bounding boxes, we can evaluate the distribution over future positions that our network infers. We interpret the sensitivty of our model, similar to \cite{interpsana}, by removing each box in a scene and evaluating the PKL. In Figure~\ref{fig:importance}, we color each box red according to the size of the PKL if we remove that box. We visualize these boxes in the global frame.

\begin{figure}[t!]
   \begin{center}
   \includegraphics[width=0.99\linewidth]{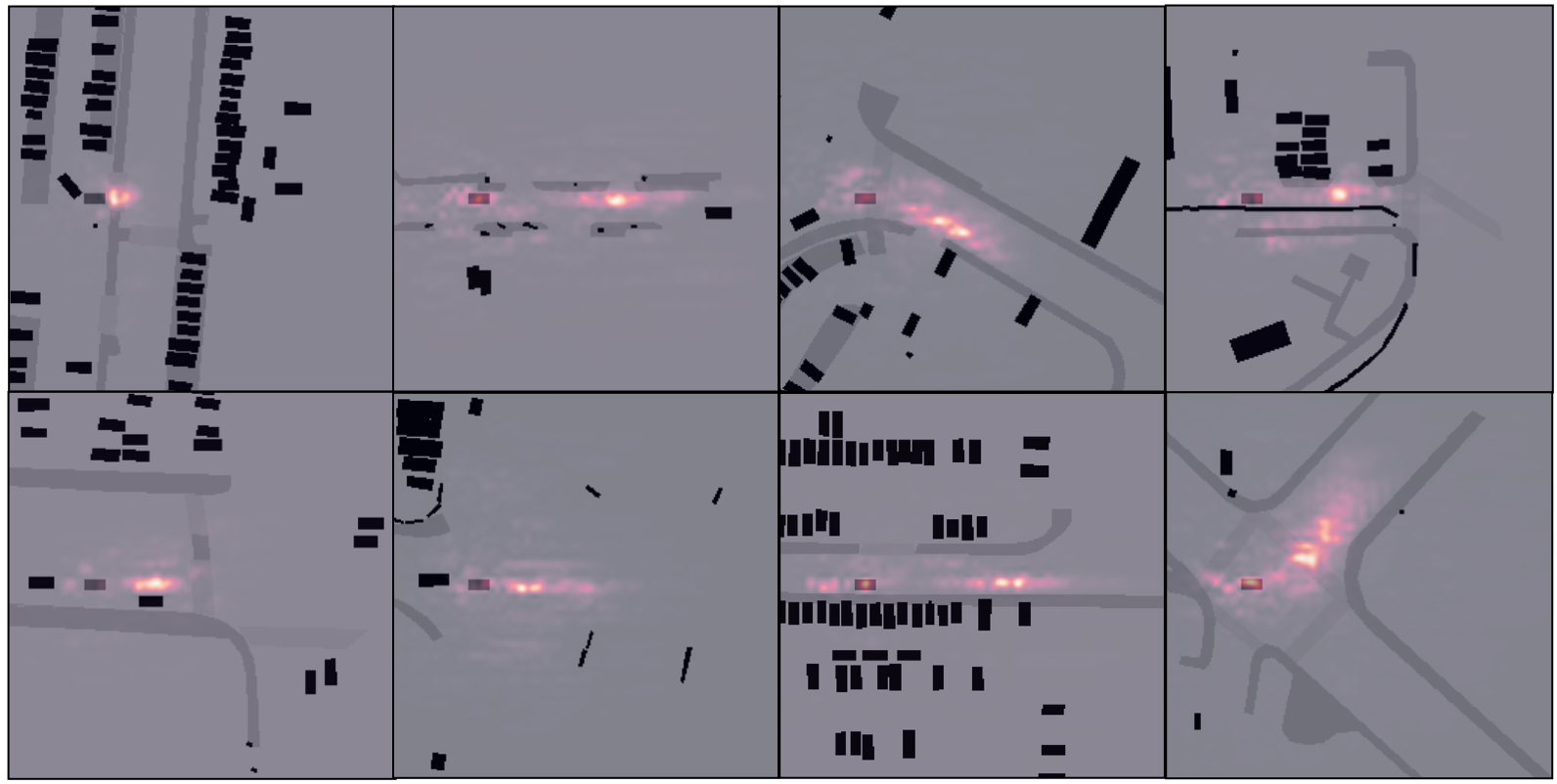}
   \end{center}
   \vspace{-3mm}
      \caption{\footnotesize{\bf False positive sensitivity} We place false positives of size 1 m by 1 m at a grid of locations for all timesteps and calculate the PKL. Regions where the false positive resulted in a higher PKL are colored red.}
   \label{fig:bad_spots}
   \vspace{-2mm}
   \end{figure}

Just as we can measure the importance of detecting every object by removing it from the scene and evaluating the PKL, we can also insert arbitrary false positives into the scene at each location $x, y$ and evaluate the PKL. This experiment measures the importance of not detecting a false positive at a certain location. As seen in Figure~\ref{fig:bad_spots}, the most dangerous locations of false positives are largely located on the current most likely path of travel for the ego vehicle.

In summary, the presented results make a strong case for planning-based metrics in evaluating perceptual models for their relevance to the downstream task.

\section{Conclusion}
\label{s:conclusion}

Our paper analyzed the current perception metrics and their relevance to the real downstream task of autonomous driving. We introduced a new planning-based metric that evaluated 3D object detections by their influence on the planner. The metric judges perception in scenes \emph{in context}, and is intrinsically responsive to multiple different error modes, which have been exploited in the past to handcraft performance metrics. 
We perform a human study, in which mechanical turkers judge the quality of different detection outputs in the same scene. Results show that even naive humans agree with our metric significantly more often than existing detection metrics, despite the fact that pre-existing metrics have been carefully designed by experts.

{\small
\bibliographystyle{ieee_fullname}
\bibliography{ref}
}

\end{document}